# Dependable Neural Networks Through Redundancy, A Comparison of Redundant Architectures

Hans Dermot Doran, Gianluca Ielpo, David Ganz
*Institute of Embedded Systems*
Zurich University of Applied Sciences
Winterthur, Switzerland
{donn, ielp, ganz} @zhaw.ch

Michael Zapke
*Industrial, Vision, Healthcare and Sciences*
Xilinx GmbH
Munich, Germany
mzapke@xilinx.com

*Abstract*— **With edge-AI finding an increasing number of real-world applications, especially in industry, the question of functionally safe applications using AI has begun to be asked. In this body of work, we explore the issue of achieving dependable operation of neural networks. We discuss the issue of dependability in general implementation terms before examining lockstep solutions. We intuit that it is not necessarily a given that two similar neural networks generate results at precisely the same time and that synchronization between the platforms will be required. We perform some preliminary measurements that may support this intuition and introduce some work in implementing lockstep neural network engines.**

## I. Introduction

### A. Motivation

Edge-AI, apart from being something of a buzzword, has obvious potential in industrial, specifically factory automation, applications. These potential applications vary from condition monitoring through to functionally safe zoning of (stationary) robot work areas. Most of these application domains are associated with some form of dependability or functional safety, for which dependability of operation is necessary, yet there has been very little concrete said about dependability in neural networks, our area of concern.

In this paper we examine the case for Field Programmable Gate Array (FPGA) and Graphics Processing Units (GPU) implementations of neural networks (NN.) We argue, based on two example architectures, that FPGA implementations appear to be affine to both tightly- and loosely-coupled lockstep circuits whereas discrete GPU implementations are more affine to loosely-coupled lockstep circuits. Whereas tightly-coupled lockstep architectures by their nature, deliver results within a few clock cycles of each other, it is not clear that loosely-coupled lockstep architectures do so. We present the results of some naïve experiments illustrating the problem. Finally, we present on-going work in the area of tightly-coupling lockstepping on an FPGA.

### B. Previous Work

Previous work in the domain of safety and neural networks is sparse although the problem has long been recognised [1]. A survey paper [2] provides an overview of literature which is less safety, in the context of IEC 61508, focused but more on the input perturbation and adversarial training/classification problem. The two main application domains are aeronautics [3] and autonomous driving [4]. The industry domains of factory automation and the like have, apart from some preliminary papers [5], received little attention so far. Integrated circuit and related manufacturers have not been idle, both Nvidia [6] and ARM [7] have released silicon or IP to support functional safety/neural network applications.

### C. Workflows and Dependability

The work-flow from a product idea to a functionally safe, or at least dependable, product is well understood and demands dependability work-packets in three distinct phases. Design: where a model of the product, including any dependability factors, is envisioned and specified. The implementation: where the specifications are synthesised into hardware and software according to a (safe and) dependable workflow and appropriately verified and validated. The run-time phase: where the integrity of operation is secured – of and by circuits previously modelled and implemented. The work flow from a neural network based product to an actual running implementation is much more complex and encompasses the choice of neural network model, number of neurons, number of layers as well as the choice of training data reflect the biases of the designer as much as any other factors [8]. The training environment takes chosen input data and runs a number of algorithms on it producing a result that the designer checks by hand, aiming for a classification rate close to the specified limit. It is an iterative process and depends as much on experience to get the targeted result as to any scientific methodology [9]. Next, some tooling is used to produce a synthesisable – in the generic meaning of the term - implementation from the output of the learning system. In a safe and dependable work flow we would expect all stages to produce inspectable results, which is clearly not currently the case. In other words, no causal chain can be guaranteed let alone validated. Finally the run-time integrity must be guaranteed, a significant challenge for such complex circuitry.

We currently concern ourselves with the last phase, that is the run-time phase and ensuring the integrity of operation.

### D. Dependable Operation and Redundancy

Redundant execution is a well-known technique that seeks to verify the integrity of an executed operation by executing that operation more than once and comparing the execution process or results of execution Figure 1. We take tightly-coupled lockstep systems, the term is not precisely defined, to mean systems where the unit of execution is executed synchronously more than once and the execution process (load-execute-store) visible on the system bus is compared. Well-known articulations of this method are lockstep processors on offer from various integrated circuit manufacturers [10, 11] whose synchronicity is fixed within typically 0 to 2 clock edges. The real-time-processors on one of the target platforms, the Xilinx ZU3EG A484, [12] can also be optionally tightly lockstepped. The term loosely-coupled lockstepping encompasses a range of circuits but is generally understood to consist of redundant circuits that can be described as asynchronous. Asynchronous execution places an additional burden on the entire circuit as the checker must be informed when checking can occur. Whilst we currently only consider 1oo2 configurations we will be working with MooN configurations which is why we use the term "voter/checker" for NN output comparison and "safety-switch-off" for a sub-circuit that checks the functional integrity of other component blocks.

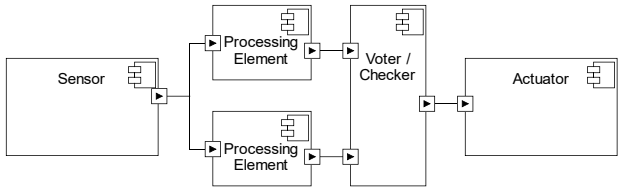

*Figure 1: Simplified Schematic of a Redundant System. The Processing Elements can be either tightly- or loosely coupled and the configuration could be either 1oo2 or 2oo2 depending on the voter implementation*

Neural networks are implemented using processing elements, these can be a CPU or, more commonly, some form of GPU or other parallel signal processing architecture. Neural networks are easily decomposed into single instruction multiple data (SIMD) kernels that can be re-grouped into threads, termed single thread multiple data (STMD.)

If we assume for the sake of simplicity the OpenCL [13] model of operations, a host, typically a CPU, will transfer these kernels to a compute device, composed of compute units in turn composed of processing elements. In a GPU, the compute unit is generally understood to be a STMD streaming processor composed of multiple SIMD cores (processing elements,) so the compute unit will decompose the threads back into individual instructions for execution across the SIMD cores. This latter process is known as fine-grained scheduling whereas the distribution of thread packages to the compute units is known as coarse-grained scheduling. These scheduling systems are native to the compute unit which are generally accessed by a driver supplied by the GPU vendor. The effect of this is that under naïve circumstances, the only influence the application implementer has over the scheduling is the order of submission of kernels to the compute unit. The driver may give the implementer the option of overriding the order of execution by stipulating in-order execution after which the coarse-grained and fine-grained scheduling is left to the compute unit. Scheduling schemes imply synchronisation points especially if the results of computation under two independent schedulers are to be compared. We therefore intuit that it is feasible that output results from redundant execution may differ in timing.

## II. REDUNDANT NEURAL NETWORK IMPLEMENTATIONS

### A. Introduction

In this body of work, we largely confine ourselves to the naïve approach, that is the use of off-the-shelf components. Given the large variation in possible approaches and circuits we model two example architectures, based on Figure 1, from which we derive the experimental setup and further work. We examine two architectures, namely Graphics Processing Units (GPU) and Field Programmable Gate Arrays (FPGA) which are well accepted to handle the computational expense and hence fulfil any real-time deadlines that may exist, as expected in factory automation systems.

### B. Graphics Processing Unit

The GPU most often used in embedded AI solutions is a collection of streaming multiprocessors each featuring an array of dedicated functional units including arithmetic, load-and-store and other assorted special functions. This array is driven by a dispatcher that issues instructions in an order determined by the on-chip schedulers. The GPU package will generally be fitted with a CPU which, according to the OpenCL model, functions as a feeder, delegating computationally expensive kernels to the GPU. This makes for a complex circuit requiring its own infrastructure circuitry, memory, I/O ports, power supply and the like. For a redundant execution model, whilst it is not impossible to develop a circuit where these infrastructure components are shared, it is not an expected use for the processor-packages which intuitively implies that modules must be architected into redundant computing units.

Our example architecture (Figure 2) thus features two GPU modules connected to an I/O source and a voter/checker circuit. Complicating, in terms of design simplicity and hence validation of causality, is the I/O circuitry – the input arrival time to the modules ought to be synchronised such that the arrival time of data is not a contributing factor to the two modules drifting temporally apart. Additional complicating factors include that the input will be routed over the CPU which delegates kernels and data to, and receives results from, the GPU. The CPU will also feed the voter/checker with this data. Under these conditions the CPU must also exhibit dependability characteristics.

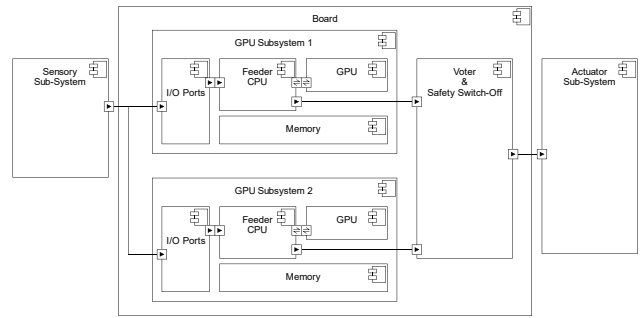

*Figure 2: Example Architecture of a GPU-Module based Loosely-Coupled Lockstepping Scheme for Dependable Operation*

### C. Field Programmable Gate Array

The field programmable gate array, in its SoC articulation, will feature programmable fabric as well as, often a plurality of, hard-core processors which can be directly connected to circuitry located in the fabric. In our example architecture two NN implementations fit into the FPGA fabric (Figure 3) along with the voter and the I/O connects directly to the fabric, a considerable simplification over the GPU-module. Isolation-design workflows are available for FPGAs so it is possible to floorplan, isolate and independently execute two NNs with defined input and output interfaces thus securing the causality chain of operation. Under this configuration the classification by redundant NNs could conceivably run without CPU intervention, thus possibly absolving CPUs of the requirement to utilise dependability features. Further analysis will determine whether this is feasible in practice.

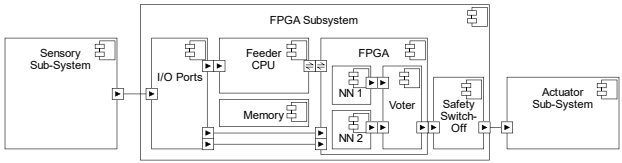

*Figure 3: Example Architecture of an FPGA-Based Tightly-Coupled Lockstepping Scheme for Dependable Operation*

*D. Experimental results*

We build two experimental systems. The first with two Nvidia Jetson TX1 development boards [14] and the second with an Avnet U96 board featuring a Xilinx MPSoC XZCU3EG Zynq device [15]. The former features a quad-core ARM Cortex A-57 (1.7 MHz.) platform in combination with a Maxwell-architecture GPU with 256 CUDA cores (998 MHz.) The latter features an ARM Cortex A-53 (1.2 GHz.) for the NN implementation we use a Xilinx Deep Learning Processing Unit (DPU – 210 MHz.) in the FPGA fabric. The relatively low speed of the DPU is because the power supply of the U96 board cannot support the DPU running at higher frequencies.

Both systems run under a nativized version of Linux, (Jetpack 4.4.1/LinuxTegra V 32.4.4 and Petalinux 2020.1/Vitis AI 1.2). In each case the CPU application is a single process, pinned to one core and scheduled with priority 99 under a FIFO scheduler.

In keeping with the naïve approach, both systems use network versions of the SSD Mobilenet V2 object detection network, pretrained by their respective vendors/community. For experimentation we used images in the range 200-799 from the EPFL dataset [16].

To measure the temporal execution-profile we loaded 500 images from the experimental dataset into the respective compute devices and executed inference 100 times for each image. The turnaround time was measured for each image inference using the chrono library.

The results of the GPU measurement are shown below (Figure 4) and show two noteworthy features. The two modules, which were not fresh out-of-the-box but had been used in other projects, display varying execution-time profiles and different outliers. The outliers themselves are presumed to come from some, as yet unexplained, CPU scheduling effect and indeed, varying the CPU frequency, where architecturally possible, affects the absolute time of these outliers. The difference between the two execution-time profiles is not understood.

The results from the FPGA measurement are shown below (Figure 5.) In this case we only measured one DPU sized to fit in the available FPGA fabric. Reasons are discussed below. Here also we see outliers that we attribute to scheduling effects, although the execution-time profile exhibits substantial kurtosis.

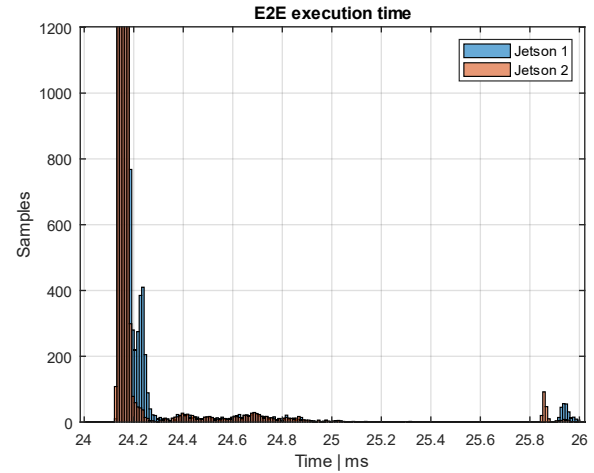

*Figure 4: Comparison of Turnaround Execution-Time Profiles for 500 Images, each Classified 100 Times, on two Similar Jetson TX1 Development Boards*

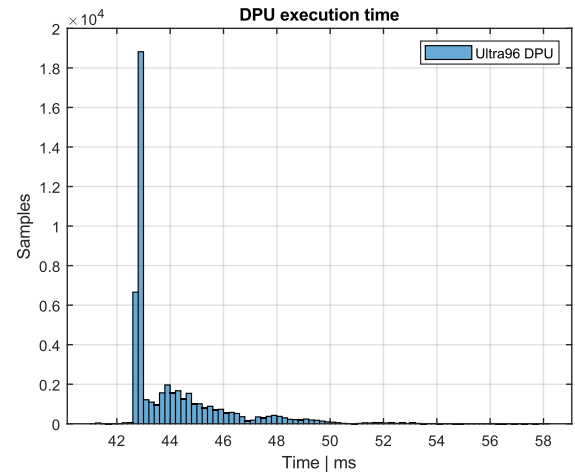

*Figure 5: Turnaround Execution-Time Profile for 500 Images each Classified 100 Times on a Xilinx FPGA Platform using Deep Learning Processing Units*

Kurtosis and bi-modality in execution-time profiles are in our experience attributable to the run-time platform, in this case primarily the operating system. It is unclear what effect, if any, the schedulers embedded in the GPU and FPGA have on execution time. On-going work will examine this point further.

*E. Discussion*

In conclusion we can say that once the outliers for both platforms are explained, a redundancy configuration for both GPU module and DPU appears feasible. The GPU configuration requires substantial engineering work in the synchronised feeding of data to the modules, the explanation and removal of the outliers and the explanation and rectification of the differing execution-time profiles. The latter two bodies of work represent work to introduce more determinism into the classification process, which we would expect a module-to-module synchronisation protocol such as IEEE1588 [17] to support.

## III. Tightly Lockstepped AI Cores

Xilinx offers two possibilities of implementing AI cores in their FPGAs, the Deep Learning Processor and FINN [18], an experimental, dataflow-orientated framework for creating quantised networks. The FINN framework appears predestined for creating custom networks and it is feasible that networks including redundancy and checkers can be built and integrated into an industry-acceptable workflow. In the context of lockstep architectures, the DPU has several attractive features that can be leveraged to produce a tightly-coupled lockstep architecture.

There are multiple configuration parameters for the DPU and the configuration/synthesizer will then pick the numbers of LUTs/DSPs that suits the synthesizer/FPGA device best and package it into an intellectual property (IP) block. This packaging brings some useful characteristics. The packaged DPU is accessible over generic well-specified bus interfaces, the number of interfaces determined in part by the number of DPU cores in the IP block. The relevant Xilinx DPU documentation does mention the possibility of having independent tasks running on DPU cores within the IP block, but the architecture also lends itself to two independent IP blocks within the same FPGA fabric. This packaging of DPU cores as an independent IP block also lends itself to standard application of the isolation design workflow, a workflow designed to produce certifiably safe and dependable implementations. By proxy, standard verification and validation workflows and regression testing frameworks both in simulation and in implementation can also be used.

The authors' research team employs a python-driven Open Verification Methodology platform [19] for simulation-based verification in the context of an automated build and test environment coupled with a continuous deployment framework for automated deployment and validation on a range of always-online run-time platforms, currently numbered at eight. This infrastructure allows the mass testing of various configurations for both experimental and verification purposes.

This reduction of the AI functionality into easily handleable IP blocks simplifies the lock-step design as now input data can be sourced from either one or n sources. The distribution of that data to the independent DPU cores can be used as a synchronising method and guarantee that the same data/code is presented to the cores at the same time. Given they are clocked at the same rate the results should be available at the same time – once the question of outliers noted above has been solved, the bus activity, data/instruction fetch and data-issue can be monitored to ensure the visible IP execution is the same for both cores. Additionally, the loosely-coupled lockstepping method of checking output data can also be employed by the voter-checker. This circuit is in the implementation stage and verification that the isolation workflow can be used with this architecture has been completed.

## IV. Conclusions and Further Work

Initial experiments measuring turnaround time on two different platforms have shown us that the interface between host and compute element, whose performance is largely dictated by the operating system, appears to introduce substantial variance into the turnaround time. We have shown that the FPGA architecture, using DPUs, has the potential to be tightly lock-stepped and also applicable in an industry acceptable dependability workflow such as the isolation workflow.

### A. Future Work

Future work is focused on optimising the DPU-based lock-step solution but several open questions including scheduler performance in the GPU/FPGA processing elements need to answered.

## References

[1] R. A. Kosiński and C. Kozłowski, "Artificial Neural Networks– Modern Systems for Safety Control," *International Journal of Occupational Safety and Ergonomics,* vol. 4, no. 3, pp. 317-332, 1998/01/01 1998.

[2] X. Huang *et al.*, "A survey of safety and trustworthiness of deep neural networks: Verification, testing, adversarial attack and defence, and interpretability," *Computer Science Review,* vol. 37, p. 100270, 2020/08/01/ 2020.

[3] A. Clavière, E. Asselin, C. Garion, and C. Pagetti, "Safety Verification of Neural Network Controlled Systems," 2020-11-05, 2020.

[4] S. Kuutti, R. Bowden, H. Joshi, R. de Temple, and S. Fallah, "Safe Deep Neural Network-Driven Autonomous Vehicles Using Software Safety Cages," Cham, 2019, pp. 150-160: Springer International Publishing.

[5] H. D. Doran and M. Reif, "Examining redundancy in the context of safe machine learning," presented at the Proceedings of the Forum for Safety & Security 2020, Virtual, 2020.

[6] Nvidia. (2020, 6 June). *NVIDIA Xavier Achieves Industry First with Expert Safety Assessment*. Available: https://blogs.nvidia.com/blog/2020/05/20/xavier-achieves-industry-first-safety-assessment/

[7] ARM. (2021, 07 June). *CPU CORTEX-A78AE*. Available: https://www.arm.com/products/silicon-ip-cpu/cortex-a/cortex-a78ae

[8] B. Kim, H. Kim, K. Kim, S. Kim, and J. Kim, "Learning Not to Learn: Training Deep Neural Networks With Biased Data," in *2019 IEEE/CVF Conference on Computer Vision and Pattern Recognition (CVPR)*, 2019, pp. 9004-9012.

[9] X. Wang and W. Cao, "Non-iterative approaches in training feed-forward neural networks and their applications," *Soft Computing,* vol. 22, no. 11, pp. 3473-3476, 2018/06/01 2018.

[10] Texas Instruments. (2021, 07 June). *TMS570LC4357*. Available: https://www.ti.com/product/TMS570LC4357

[11] Renesas. (2021, 07 June). *RH850 Automotive MCUs*. Available: https://www.renesas.com/us/en/products/microcontrollers-microprocessors/rh850-automotive-mcus

[12] Xilinx. (2021, 07 June). *Zynq UltraScale+ MPSoC*. Available: https://www.xilinx.com/products/silicon-devices/soc/zynq-ultrascale-mpsoc.html

[13] Khronos Group. (2021, 07 June). *OpenCL: Open Standard for Parallel Programming of Heterogeneous Systems*. Available: https://www.khronos.org/opencl/

[14] Nvidia. (2021, 07 June). *Jetson TX1 Developer Kit*. Available: https://developer.nvidia.com/embedded/jetson-tx1-developer-kit

[15] Xilinx. (2021, 07 June). *Avnet Ultra96-V2*. Available: https://www.xilinx.com/products/boards-and-kits/1-vad4rl.html

[16] T. Bagautdinov. (2021). *RGB-D Pedestrian Dataset*. Available: https://www.epfl.ch/labs/cvlab/data/data-rgbd-pedestrian/

[17] IEEE Standard for a Precision Clock Synchronization Protocol for Networked Measurement and Control Systems, 2019.

[18] Xilinx. (2020, 08 June). *FINN Documents*. Available: https://finn.readthedocs.io/en/latest/

[19] PotentialVentures. (2020, April 2). *Introduction* Available: https://cocotb.readthedocs.io/en/latest/introduction.html